\documentclass[11pt,a4paper]{article}
\usepackage[hyperref]{acl2018}
\usepackage{times}
\usepackage{latexsym}
\usepackage{graphicx}
\usepackage{enumitem}
\usepackage{amsmath}
\usepackage{amssymb}
\usepackage{array,multirow}

\usepackage{etoolbox}
\makeatletter
\patchcmd\@combinedblfloats{\box\@outputbox}{\unvbox\@outputbox}{}{\errmessage{\noexpand patch failed}}
\makeatother
\newcolumntype{?}{!{\vrule width 1pt}}
\newcolumntype{@}{!{\vrule width 1.5pt}}
\usepackage[T1]{fontenc}
\usepackage{inconsolata}

\usepackage{booktabs}

\usepackage{url}

\aclfinalcopy % Uncomment this line for the final submission
\def\aclpaperid{1253} %  Enter the acl Paper ID here

\title{Multi-Relational Question Answering from Narratives:\\ Machine Reading and Reasoning in Simulated Worlds}

\author{
	Igor Labutov \quad Bishan Yang \\
         Machine Learning Dept. \\ Carnegie Mellon University \\ Pittsburgh, PA 15213\\
	{\tt ilabutov@cs.cmu.edu}\\{\tt bishan@cs.cmu.edu} \And
	Anusha Prakash\\
         Language Technologies Inst.\\ Carnegie Mellon University \\ Pittsburgh, PA 15213\\
	{\tt anushap@andrew.cmu.edu } \And
	Amos Azaria \\
         Computer Science Dept.\\ Ariel University\\ Israel\\
	{\tt amos.azaria@ariel.ac.il}
}

\date{}

\begin{document}
%\twocolumn
\maketitle
\begin{abstract}
Question Answering (QA), as a research field, has primarily focused on either knowledge bases (KBs) or free text as a source of knowledge. These two sources have historically shaped the kinds of questions that are asked over these sources, and the methods developed to answer them. 
%Questions over knowledge bases can generally be highly compositional, but are constrained to a finite number of relations defined in the KB. Questions over free text, on the other hand, are unrestricted in terms of relations, but generally constrain the answer to be a single entity extracted from the text. 
In this work, we look towards a practical use-case of \textit{QA over user-instructed knowledge} that uniquely combines elements of both \textit{structured QA} over knowledge bases, and \textit{unstructured QA} over narrative, introducing the task of \textit{multi-relational QA over personal narrative}. As a first step towards this goal, we make three key contributions: (i) we generate and release {\sc TextWorldsQA}, a set of five diverse datasets, where each dataset contains dynamic narrative that describes entities and relations in a simulated world, paired with variably compositional questions over that knowledge, (ii) we perform a thorough evaluation and analysis of several state-of-the-art QA models and their variants at this task, and (iii) we release a lightweight Python-based framework we call {\sc TextWorlds} for easily generating arbitrary additional worlds and narrative, with the goal of allowing the community to create and share a growing collection of diverse worlds as a test-bed for this task.

\end{abstract}

\section{Introduction}
\begin{figure}
\label{fig:cartoon}
\includegraphics[width=8cm]{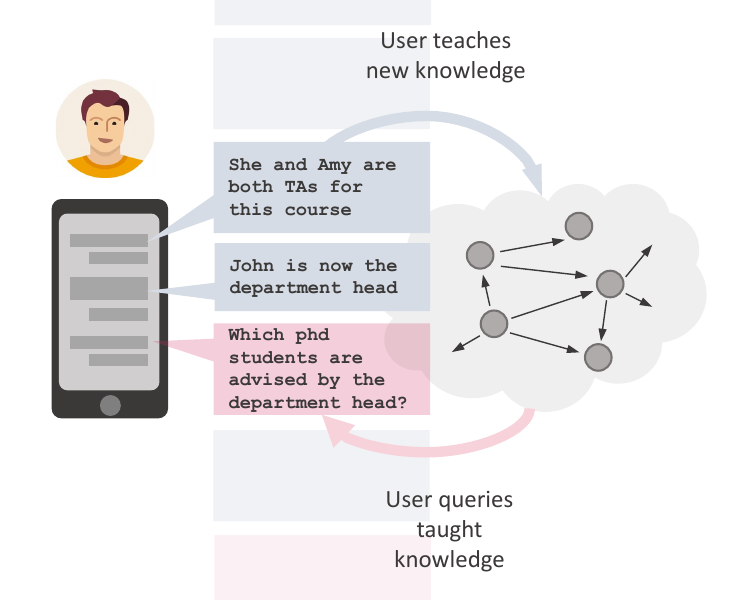}
\centering
\caption{Illustration of our task: relational question answering from dynamic knowledge expressed via personal narrative}
\end{figure}
\begin{figure*}[ht]
\label{fig:text_examples}
\includegraphics[width=0.9\textwidth]{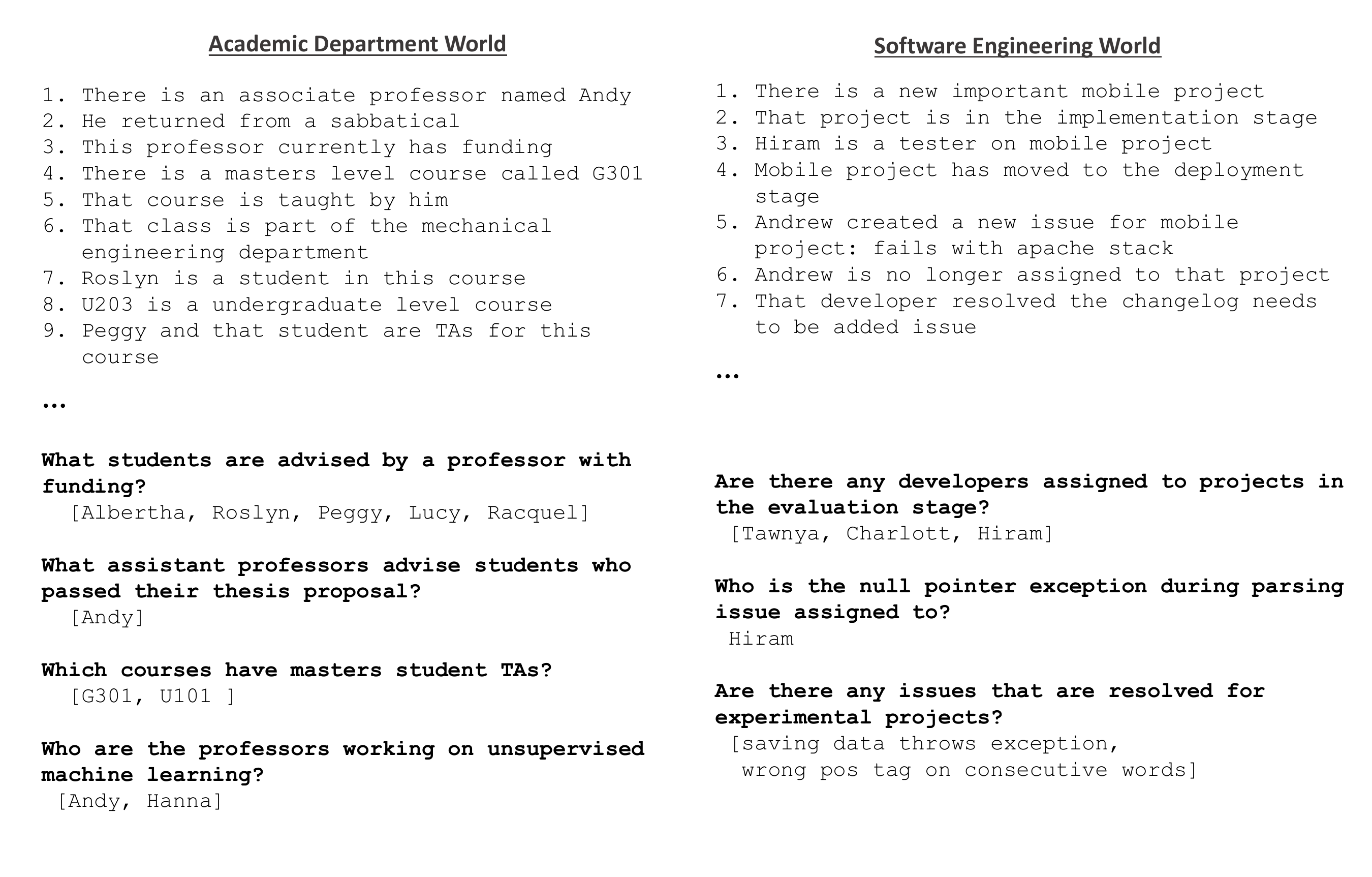}
\centering
\caption{Illustrative snippets from two sample worlds. We aim to generate natural-sounding first-person narratives from five diverse worlds, covering a range of different events, entities and relations.}
\end{figure*}
Personal devices that interact with users via natural language conversation are becoming ubiquitous (e.g., Siri, Alexa), however, very little of that conversation today allows the user to teach, and then query, new knowledge. Most of the focus in these personal devices has been on Question Answering (QA) over general world-knowledge (e.g., \textit{``who was the president in 1980''} or \textit{``how many ounces are in a cup''}).
These devices open a new and exciting possibility of enabling end-users to teach machines in natural language, e.g., by expressing the state of their personal world to its virtual assistant (e.g., via narrative about people and events in that user's life) and enabling the user to ask questions over that personal knowledge (e.g., \textit{``which engineers in the QC team were involved in the last meeting with the director?''}).

This type of questions highlight a unique blend of two conventional streams of research in Question Answering (QA) -- QA over \textit{structured} sources such as knowledge bases (KBs), and QA over \textit{unstructured} sources such as free text. This blend is a natural consequence of our problem setting: (i) users may choose to express rich relational knowledge about their world, in turn enabling them to pose complex \textbf{compositional} queries (e.g., \textit{``all CS undergrads who took my class last semester''}), while at the same time (ii) personal knowledge generally evolves through time and has an open and growing set of relations, making natural language the only practical interface for creating and maintaining that knowledge by non-expert users. In short, the task that we address in this work is: \textbf{multi-relational question answering from dynamic knowledge expressed via narrative}.

Although we hypothesize that question-answering over personal knowledge of this sort is ubiquitous (e.g., between a professor and their administrative assistant, or even if just in the user's head), such interactions are rarely recorded, presenting a significant practical challenge to collecting a sufficiently large real-world dataset of this type. At the same time, we hypothesize that the technical challenges involved in developing models for relational question answering from narrative would not be fundamentally impacted if addressed via sufficiently rich, but controlled simulated narratives. Such simulations also offer the advantage of enabling us to directly experiment with stories and queries of different complexity, potentially offering additional insight into the fundamental challenges of this task.

While our problem setting blends the problems of relational question answering over knowledge bases and question answering over text, our hypothesis is that end-to-end QA models may learn to answer such multisentential relational queries, without relying on an intermediate knowledge base representation. In this work, we conduct an extensive evaluation of a set of state-of-the-art end-to-end QA models on our task and analyze their results.

\section{Related Work}
Question answering has been mainly studied in two different settings: KB-based and text-based. KB-based QA mostly focuses on parsing questions to logical forms~\cite{zelle1996learning,zettlemoyer2012learning,berant2013semantic, kwiatkowski2013scaling, yih2015semantic} in order to better retrieve answer candidates from a knowledge base. Text-based QA aims to directly answer questions from the input text. This includes works on early information retrieval-based methods~\cite{banko2002askmsr,ahn2004using} and methods that build on extracted structured representations from both the question and the input text~\cite{sachan2015learning,sachan2016machine,khot2017answering,khashabi2018question}. Although these structured presentations make reasoning more effective, they rely on sophisticated NLP pipelines and suffer from error propagation. More recently, end-to-end neural architectures have been successfully applied to text-based QA, including Memory-augmented neural networks~\cite{sukhbaatar2015end, miller2016key, kumar2016ask} and attention-based neural networks~\cite{hermann2015teaching,chen2016thorough,kadlec2016text,dhingra2016gated,xiong2016dynamic,seo2016bidirectional,chen2017reading}. In this work, we focus on QA over text (where the text is generated from a supporting KB) and evaluate several state-of-the-art memory-augmented and attention-based neural architectures on our QA task. In addition, we consider a sequence-to-sequence model baseline~\cite{bahdanau2014neural}, which has been widely used in dialog~\cite{vinyals2015neural,ghazvininejad2017knowledge} and recently been applied to generating answer values from Wikidata~\cite{hewlett2016wikireading}.

There are numerous datasets available for evaluating the capabilities of QA systems. For example, MCTest~\cite{richardson2013mctest} contains comprehension questions for fictional stories. Allen AI Science Challenge~\cite{clark2015elementary} contains science questions that can be answered with knowledge from text books. {\sc RACE}~\cite{lai2017race} is an English exam dataset for middle and high school Chinese students. {\sc MultiRC}~\cite{daniel2018} is a dataset that focuses on evaluating multi-sentence reasoning skills. These datasets all require humans to carefully design multiple-choice questions and answers, so that certain aspects of the comprehension and reasoning capabilities are properly evaluated. As a result, it is difficult to collect them at scale. Furthermore, as the knowledge required for answering each question is not clearly specified in these datasets, it can be hard to identify the limitations of QA systems and propose improvements.  

\newcite{weston2015towards} proposes to use synthetic QA tasks (the {\sc bAbI} dataset) to better understand the limitations of QA systems. {\sc bAbI} builds on a simulated physical world similar to interactive fiction~\cite{montfort2005twisty} with simple objects and relations and includes 20 different reasoning tasks. Various types of end-to-end neural networks~\cite{sukhbaatar2015end,lee2015reasoning,peng2015towards} have demonstrated promising accuracies on this dataset. However, the performance can hardly translate to real-world QA datasets, as {\sc bAbI} uses a small vocabulary (150 words) and short sentences with limited language variations (e.g., nesting sentences, coreference). A more sophisticated QA dataset with a supporting KB is {\sc WikiMovies}~\cite{miller2016key}, which contains ~100k questions about movies, each of them is answerable by using either a KB or a Wikipedia article. However, {\sc WikiMovies} is highly domain-specific, and similar to {\sc bAbI}, the questions are designed to be in simple forms with little compositionality and hence limit the difficulty level of the tasks. 

Our dataset differs in the above datasets in that (i) it contains five different realistic domains permitting cross-domain evaluation to test the ability of models to generalize beyond a fixed set of KB relations, (ii) it exhibits rich referring expressions and linguistic variations (vocabulary much larger than the \textsc{bAbI} dataset), (iii) questions in our dataset are designed to be deeply compositional and can cover multiple relations mentioned across multiple sentences.

Other large-scale QA datasets include Cloze-style datasets such as CNN/Daily Mail~\cite{hermann2015teaching}, Children's Book Test~\cite{hill2015goldilocks}, and Who Did What~\cite{onishi2016did}; datasets with answers being spans in the document, such as SQuAD~\cite{rajpurkar2016squad}, NewsQA~\cite{trischler2016newsqa}, and TriviaQA~\cite{joshi2017triviaqa}; and datasets with human generated answers, for instance, MS MARCO~\cite{nguyen2016ms} and SearchQA~\cite{dunn2017searchqa}. One common drawback of these datasets is the difficulty in accessing a system's capability of integrating information across a document context.~\newcite{kovcisky2017narrativeqa} recently emphasized this issue and proposed NarrativeQA, a dataset of fictional stories with questions that reflect the complexity of narratives: characters, events, and evolving relations. Our dataset contains similar narrative elements, but it is created with a supporting KB and hence it is easier to analyze and interpret results in a controlled setting. 

\section{\textsc{TextWorlds: }Simulated Worlds for Multi-Relational QA from Narratives }
In this work, we synthesize narratives in five diverse worlds, each containing a thousand narratives and where each narrative describes the evolution of a simulated user's world from a first-person perspective. In each narrative, the simulated user may introduce new knowledge, update existing knowledge or express a state change (e.g., \textit{``Homework 3 is now due on Friday''} or \textit{``Samantha passed her thesis defense''}). Each narrative is interleaved with questions about the current state of the world, and questions range in complexity depending on the amount of knowledge that needs to be integrated to answer them. This allows us to benchmark a range of QA models at their ability to answer questions that require different extents of relational reasoning to be answered.

The set of worlds that we simulate as part of this work are as follows:
\begin{enumerate}[leftmargin=*]
    \item {\sc Meeting world:} This world describes situations related to professional meetings, e.g., meetings being set/cancelled, people attending meetings, topics of meetings.
    \item {\sc Homework world:} This world describes situations from the first-person perspective of a student, e.g., courses taken, assignments in different courses, deadlines of assignments.
    \item {\sc Software engineering world:} This world describes situations from the first-person perspective of a software development manager, e.g., task assignment to different project team members, stages of software development, bug tickets. 
    \item {\sc Academic department world:} This world describes situations from the first-person perspective of a professor, e.g., teaching assignments, faculty going/returning from sabbaticals, students from different departments taking/dropping courses.
    \item {\sc Shopping world:} This world describes situations about a person shopping for various occasions, e.g., adding items to a shopping list, purchasing items at different stores, noting where items are on sale.
\end{enumerate}
\begin{table}
\begin{footnotesize}
\begin{center}
\begin{tabular}{ll}
\toprule
Statistics & Value\\
\midrule
\# of total stories & 5,000\\
\# of total questions & 1,207,022\\
Avg. \# of entity mentions (per story) & 217.4\\
Avg. \# of correct answers (per question) & 8.7\\
Avg. \# of statements in stories & 100\\
Avg. \# of tok. in stories & 837.5\\
Avg. \# of tok. in questions & 8.9\\
Avg. \# of tok. in answers & 1.5\\
Vocabulary size (tok.) & 1,994\\
Vocabulary size (entity) & 10,793\\
\bottomrule
\end{tabular}
\end{center}
\caption{{\sc TextWorldsQA} dataset statistics}
\label{data-stats}
\end{footnotesize}
\end{table}
\begin{table*}[ht]
\small
\begin{center}
\begin{tabular}{|l@c@c|c|c|}
\hline \bf Dataset & \multicolumn{4}{c|}{\bf Questions } \\ 
\cline{2-5}   \bf & Single Entity/Relation & \multicolumn{3}{c|}{Multiple entities}\\ \cline{3-5}
 & & Single relation & Two relations & Three relations \\
\cline{1-5}
\sc Meeting & 57,590 (41.16\%) & 46,373 (33.14\%) & 30,391 (21.72\%) & 5,569 (3.98\%) \\
\sc Homework & 45,523 (24.10\%) & 17,964 (9.51\%) & 93,669 (49.59\%) & 31,743 (16.80\%) \\
\sc Software & 47,565 (20.59\%) & 51,302 (22.20\%) & 66,026 (28.58\%) & 66,150 (28.63\%) \\
\sc Academic & 46,965 (24.81\%) & 54,581 (28.83\%) & 57,814 (30.53\%) & 29,982 (15.83\%) \\
\sc Shopping  & 111,522 (26.25\%) & 119,890 (28.22\%) & 107,418 (25.29\%) & 85,982 (20.24\%) \\
\hline
\sc \bf All & 309,165 (26.33\%) & 290,110 (24.71\%) & 355,318 (30.27\%) & 219,426 (18.69\%) \\ \hline
\end{tabular}
\end{center}
\caption{\label{data-stats-by-type} Dataset statistics by question type. }
\end{table*}

\subsection{Narrative}
Each world is represented by a set of entities $\mathcal{E}$ and a set of unary, binary or ternary relations $\mathcal{R}$. Formally, a single step in one simulation of a world involves a combination of instantiating new entities and defining new (or mutating existing) relations between entities. Practically, we implement each world as a collection of classes and methods, with each step of the simulation creating or mutating class instances by sampling entities and methods on those entities. By design, these classes and methods are easy to extend, to either enrich existing worlds or create new ones. Each simulation step is then expressed as a natural language statement, which is added to the narrative. In the process of generating a natural language expression, we employ a rich mechanism for generating anaphora, such as \textit{``meeting with John about the performance review''} and \textit{``meeting that I last added''}, in addition to simple pronoun references. This allows us to generate more natural and flowing narratives. These references are generated and composed automatically by the underlying \textsc{TextWorlds} framework, significantly reducing the effort needed to build new worlds. Furthermore, all generated stories also provide additional annotation that maps all entities to underlying gold-standard KB ids, allowing to perform experiments that provide models with different degrees of access to the ``simulation oracle''.

We generate 1,000 narratives within each world, where each narrative consists of 100 sentences, plus up to 300 questions interleaved randomly within the narrative. See Figure \ref{fig:text_examples} for two example narratives. Each story in a given world samples its entities from a large general pool of entity names collected from the web (e.g., \textit{people names}, \textit{university names}). Although some entities do overlap between stories, each story in a given world contains a unique flow of events and entities involved in those events. See Table~\ref{data-stats} for the data statistics.

\subsection{Questions}
\label{sec:questions}
Formally, questions are queries over the knowledge-base in the state defined up to the point when the question is asked in the narrative. In the narrative, the questions are expressed in natural language, employing the same anaphora mechanism used in generating the narrative (e.g., \textit{``who is attending the last meeting I added?''}). 

We categorize generated questions into four types, reflecting the number and types of facts required to answer them; questions that require more facts to answer are typically more compositional in nature. We categorize each question in our dataset into one of the following four categories:

\vspace{2mm}
\noindent \textbf{Single Entity/Single Relation}
Answers to these questions are a single entity, e.g. \textit{``what is John's email address?''}, or expressed in lambda-calculus notation:
$$
\lambda x.\texttt{EmailAddress}(\texttt{John}, x)
$$
The answers to these questions are found in a single sentence in the narrative, although it is possible that the answer may change through the course of the narrative (e.g., \textit{``John's new office is GHC122''}).

\vspace{2mm}
\noindent \textbf{Multi-Entity/Single Relation}
Answers to these questions can be multiple entities but involve a single relation, e.g., \textit{``Who is enrolled in the Math class?''}, or expressed in lambda calculus notation:
\begin{align}
\lambda x. &\texttt{TakingClass}(x, \texttt{Math}) \nonumber
\end{align}
Unlike the previous category, answers to these questions can be sets of entities. 

\vspace{2mm}
\noindent \textbf{Multi-Entity/Two Relations}
Answers to these questions can be multiple entities and involve two relations, e.g., \textit{``Who is enrolled in courses that I am teaching?''}, or expressed in lambda calculus:
\begin{align}
\lambda x. \exists y.&\texttt{EnrolledInClass}(x, y)\nonumber \\
&\wedge \texttt{CourseTaughtByMe}(y) \nonumber
\end{align}

\vspace{2mm}
\noindent \textbf{Multi-Entity/Three Relations}
Answers to these questions can be multiple entities and involve three relations, e.g., \textit{``Which undergraduates are enrolled in courses that I am teaching?''}, or expressed in lambda calculus notation:
\begin{align}
\lambda x. \exists y.&\texttt{EnrolledInClass}(x, y)\nonumber \\
&\wedge \texttt{CourseTaughtByMe}(y) \nonumber \\
&\wedge \texttt{Undergrad}(x) \nonumber
\end{align}
In the data that we generate, answers to questions are always sets of spans in the narrative (the reason for this constraint is for easier evaluation of several existing machine-reading models; this assumption can easily be relaxed in the simulation). In all of our evaluations, we will partition our results by one of the four question categories listed above, which we hypothesize correlates with the difficulty of a question.

\section{Methods}
We develop several baselines for our QA task, including a logistic regression model and four different neural network models: Seq2Seq~\cite{bahdanau2014neural}, MemN2N~\cite{sukhbaatar2015end}, BiDAF~\cite{seo2016bidirectional}, and DrQA~\cite{chen2017reading}. These models generate answers in different ways, e.g., predicting a single entity, predicting spans of text, or generating answer sequences. Therefore, we implement two experimental settings: {\sc Entity} and {\sc Raw}. In the {\sc Entity} setting, given a question and a story, we treat all the entity spans in the story as candidate answers, and the prediction task becomes a classification problem. In the {\sc Raw} setting, a model needs to predict the answer spans. For logistic regression and MemN2N, we adopt the {\sc Entity} setting as they are naturally classification models. This ideally provides an upper bound on the performance when considering answer candidate generation. For all the other models, we can apply the {\sc Raw} setting. %For the Seq2Seq model we adopt the {\sc Raw} setting in the \textit{within-world} evaluation setting, while using the {\sc Entity} setting in the \textit{across-world} evaluation setting. For BIDAF-M and DrQA-M, we adopt the {\sc Raw} setting.

\subsection{Logistic Regression}
The logistic regression baseline predicts the likelihood of an answer candidate being a true answer. For each answer candidate $e$ and a given question, we extract the following features: (1) The frequency of $e$ in the story; (2) The number of words within $e$; (3) Unigrams and bigrams within $e$; (4) Each non-stop question word combined with each non-stop word within $e$; (5) The average minimum distance between each non-stop question word and $e$ in the story; (6) The common words (excluding stop words) between the question and the text surrounding of $e$ (within a window of 10 words); (7) Sum of the frequencies of the common words to the left of $e$, to the right $e$, and both. These features are designed to help the model pick the correct answer spans. They have shown to be effective for answer prediction in previous work~\cite{chen2016thorough,rajpurkar2016squad}. 

We associate each answer candidate with a binary label indicating whether it is a true answer. We train a logistic regression classifier to produce a probability score for each answer candidate. During test, we search for an optimal threshold that maximizes the F1 performance on the validation data. During training, we optimize the cross-entropy loss using Adam~\cite{kingma2014adam} with an initial learning rate of 0.01. We use a batch size of $10,000$ and train with 5 epochs. Training takes roughly 10 minutes for each domain on a Titan X GPU.

\subsection{Seq2Seq}
The seq2seq model is based on the sequence to sequence model presented in \cite{bahdanau2014neural}, which includes an attention model. Bahdanau et al. \cite{bahdanau2014neural} have used this model to build a neural based machine translation performing at the state-of-the-art. We adopt this model to fit our own domain by including a preprocessing step in which all statements are concatenated with a dedicated token, while eliminating all previously asked questions, and the current question is added at the end of the list of statements. The answers are treated as a sequence of words. We use word embeddings \cite{zou2013bilingual}, as it was shown to improve accuracy. We use $3$ GRU \cite{cho2014learning} connected layers, each with a capacity of $256$. Our batch size was set to 16. We use gradient descent with an initial learning rate of $0.5$ and a decay factor of $0.99$, iterating on the data for $50,000$ steps ($5$ epochs). The training process for each domain took approximately $48$ hours on a Titan X GPU. 
%In order to allow generalization across different domains, we replace entities appearing in each story with an id that correlates to their appearance order. That is, the first entity that appears in the story is replaced with the phrase ENTITY1, the second entity with ENTITY2, etc. After the model outputs its prediction, the entity ids are converted back to the entity phrase.

\subsection{MemN2N}
End-To-End Memory Network (MemN2N) is a neural architecture that encodes both long-term and short-term context into a memory and iteratively reads from the memory (i.e., multiple hops) relevant information to answer a question~\cite{sukhbaatar2015end}. It has been shown to be effective for a variety of question answering tasks~\cite{weston2015towards, sukhbaatar2015end,hill2015goldilocks}.

In this work, we directly apply MemN2N to our task with a small modification. Originally, MemN2N was designed to produce a single answer for a question, so at the prediction layer, it uses softmax to select the best answer from the answer candidates. In order to account for multiple answers for a given question, we modify the prediction layer to apply the logistic function and optimize the cross entropy loss instead. For training, we use the parameter setting as in a publicly available MemN2N~\footnote{\url{https://github.com/domluna/memn2n}} except that we set the embedding size to $300$ instead of $20$. We train the model for 100 epochs and it takes about 2 hours for each domain on a Titan X GPU.

\begin{table*}
\begin{footnotesize}
\begin{center}
\begin{tabular}{lcccccc}
\toprule
\textbf{Within-World} & \sc Meeting & \sc Homework & \sc Software & \sc Department & \sc Shopping & Avg. F1\\
\midrule
Logistic Regression & 50.1 & 55.7 & 60.9 & 55.9 & 61.1 & 56.7\\
Seq2Seq & 22.5 & 32.6 & 16.7 & 39.1 & 31.5 &  28.5\\
MemN2N & 55.4 & 46.6 & 69.5 & 67.3 & 46.3 & 57.0 \\
BiDAF-M & 81.8 & 76.9 & 68.4 & 68.2 & 68.7 & 72.8\\
DrQA-M & 81.2 & 83.6 & 79.1 & 76.4 & 76.5 & 79.4\\
\bottomrule
\toprule
\textbf{Cross-World} & \sc Meeting & \sc Homework & \sc Software & \sc Department & \sc Shopping & Avg. F1\\
\midrule
Logistic Regression & 9.0 & 9.1 & 11.1 & 9.9 & 7.2 & 9.3\\
Seq2Seq & 8.8 & 3.5 & 1.9 & 5.4 & 2.6 & 4.5\\
MemN2N & 23.6 & 2.9 & 4.7 & 14.6 & 0.07 & 9.2\\
BiDAF-M & 34.0 & 6.9 & 16.1 & 22.2 & 3.9 & 16.6\\
DrQA-M & 46.5 & 12.2 & 23.1 & 28.5 & 9.3 & 23.9\\
\bottomrule
\end{tabular}
\end{center}
\caption{$F_1$ scores for different baselines evaluated on both \textit{within-world} and \textit{across-world} settings.}
\label{macro-f1-results}
\end{footnotesize}
\end{table*}

\subsection{BiDAF-M}
BiDAF (Bidirectional Attention Flow Networks)~\cite{seo2016bidirectional} is one of the top-performing models on the span-based question answering dataset SQuAD~\cite{rajpurkar2016squad}. We reimplement BiDAF with simplified parameterizations and change the prediction layer so that it can predict multiple answer spans.  

Specifically, we encode the input story $\{x_1,...,x_T\}$ and a given question $\{q_1,...,q_J\}$ at the character level and the word level, where the character level uses CNNs and the word level uses pre-trained word vectors. The concatenation of the character and word embeddings are passed to a bidirectional LSTM to produce a contextual embedding for each word in the story context and in the question. Then, we apply the same bidirectional attention flow layer to model the interactions between the context and question embeddings, producing question-aware feature vectors for each word in the context, denoted as ${\bf G}\in \mathbb{R}^{d_g\times T}$. ${\bf G}$ is then fed into a bidirectional LSTM layer to obtain a feature matrix ${\bf M}_1\in \mathbb{R}^{d_1\times T}$ for predicting the start offset of the answer span, and ${\bf M}_1$ is then passed into another bidirectional LSTM layer to obtain a feature matrix ${\bf M}_2\in \mathbb{R}^{d_2\times T}$ for predicting the end offset of the answer span. We then compute two probability scores for each word $i$ in the narrative: ${\bf p}^{start}=\text{sigmoid}({\bf w}_1^T[{\bf G};{\bf M}_1])$ and ${\bf p}^{end}=\text{sigmoid}({\bf w}_2^T[{\bf G};{\bf M}_1;{\bf M}_2])$, where ${\bf w}_1$ and ${\bf w}_2$ are trainable weights. The training objective is simply the sum of cross-entropy losses for predicting the start and end indices.

We use 50 1D filters for CNN character embedding, each with a width of 5. The word embedding size is 300 and the hidden dimension for LSTMs is 128. For optimization, we use Adam~\cite{kingma2014adam} with an initial learning rate of 0.001, and use a minibatch size of 32 for 15 epochs. The training process takes roughly 20 hours for each domain on a Titan X GPU.

\subsection{DrQA-M}
DrQA~\cite{chen2017reading} is an open-domain QA system that has demonstrated strong performance on multiple QA datasets. We modify the Document Reader component of DrQA and implement it in a similar framework as BiDAF-M for fair comparisons. First, we employ the same character-level and word-level encoding layers to both the input story and a given question. We then use the concatenation of the character and word embeddings as the final embeddings for words in the story and in the question. We compute the aligned question embedding~\cite{chen2017reading} as a feature vector for each word in the story and concatenate it with the story word embedding and pass it into a bidirectional LSTM to obtain the contextual embeddings ${\bf E}\in \mathbb{R}^{d\times T}$ for words in the story. Another bidirectional LSTM is used to obtain the contextual embeddings for the question, and self-attention is used to compress them into one single vector ${\bf q}\in \mathbb{R}^{d}$. The final prediction layer uses a bilinear term to compute scores for predicting the start offset: ${\bf p}^{start}=\text{sigmoid}({\bf q}^T{\bf W}_1{\bf E})$ and another bilinear term for predicting the end offset: ${\bf p}^{end}=\text{sigmoid}({\bf q}^T{\bf W}_2{\bf E})$, where ${\bf W}_1$ and ${\bf W}_2$ are trainable weights. The training loss is the same as in BiDAF-M, and we use the same parameter setting. Training takes roughly 10 hours for each domain on a Titan X GPU.

\section{Experiments}
We use two evaluation settings for measuring performance at this task: \textit{within-world} and \textit{across-world}. In the \textit{within-world} evaluation setting, we test on the same world that the model was trained on. We then compute the precision, recall and $F_1$ for each question and report the macro-average F1 score for questions in each world. In the \textit{across-world} evaluation setting, the model is trained on four out of the five worlds, and tested on the remaining world. The \textit{across-world} regime is obviously more challenging, as it requires the model to be able to learn to generalize to unseen relations and vocabulary. We consider the \textit{across-world} evaluation setting to be the main evaluation criteria for any future models used on this dataset, as it mimics the practical requirement of any QA system used in personal assistants: it has to be able to answer questions on any new domain the user introduces to the system.

\subsection{Results}
\begin{figure}
\label{fig:breakdown-graph}
\includegraphics[width=8.2cm]{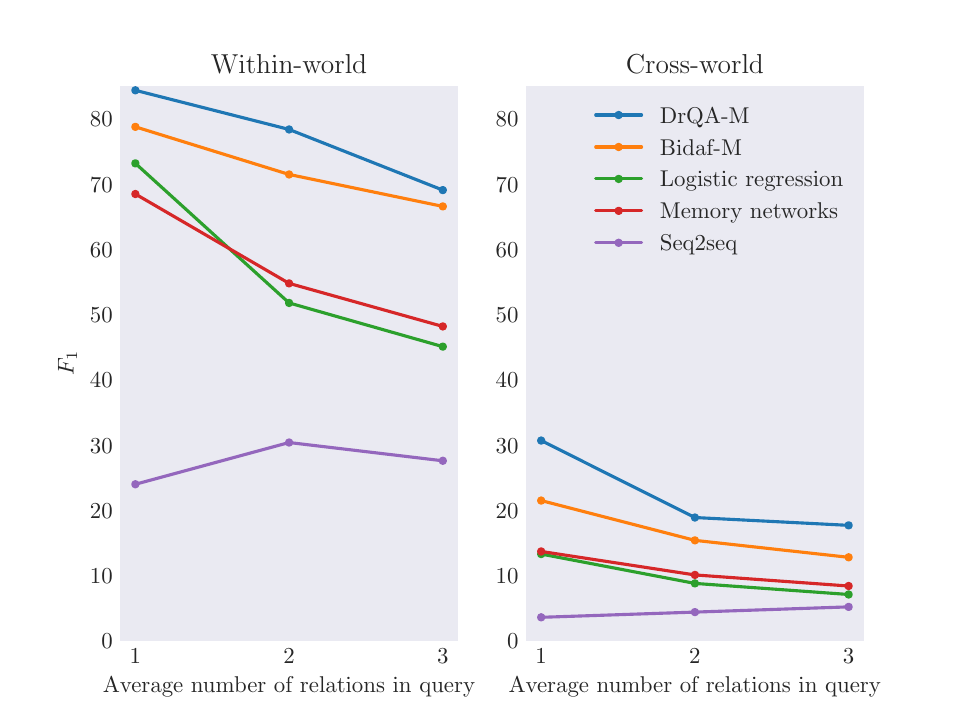}
\centering
\caption{$F_1$ score breakdown based on the number of relations involved in the questions.}
\end{figure}

We draw several important observations from our results. First, we observe that more compositional questions (i.e., those that integrate multiple relations) are more challenging for most models - as all models (except Seq2seq) decrease in performance with the number of relations composed in a question (Figure \ref{fig:breakdown-graph}). This can be in part explained by the fact that more composition questions are typically longer, and also require the model to integrate more sources of information in the narrative in order to answer them. One surprising observation from our results is that the performance on questions that ask about a single relation and have only a single answer is lower than questions that ask about a single relation but that can have multiple answers (see detailed results in the Appendix). This is in part because questions that can have multiple answers typically have canonical entities as answers (e.g., person's name), and these entities generally repeat in the text, making it easier for the model to find the correct answer. 

%The reason for this is the different nature of the single-entity/single-relation questions. As demonstrated by our example single-entity/single-relation question in Section \ref{sec:questions}, such questions are typically queries for properties of entities (e.g., email address, office location, etc). By their nature, answers to these questions are found only in one sentence. This is in contrast to questions that can have multiple entities as their answer, making it possible for the model to get ``partial credit'' by getting some of the entities right. In addition, answers to such questions in our dataset (e.g., \textit{``which students are in my class?''}) will typically be entity names, which by their nature appear multiple times in the narrative, making it more likely for the model to pick up a more frequent answer as correct in general.

Table~\ref{macro-f1-results} reports the overall (macro-average) F1 scores for different baselines. We can see that BiDAF-M and DrQA-M perform surprisingly well in the \textit{within-world} evaluation even though they do not use any entity span information. In particular, DrQA-M outperforms BiDAF-M which suggests that modeling question-context interactions using simple bilinear terms have advantages over using more complex bidirectional attention flows. The lower performance of MemN2N suggests that its effectiveness on the {\sc bAbI} dataset does not directly transfer to our dataset. Note that the original MemN2N architecture uses simple bag-of-words and position encoding for sentences. This may work well on dataset with a simple vocabulary, for example, MemN2N performs the best in the {\sc Software} world as the {\sc Software} world has a smaller vocabulary compared to other worlds. In general, we believe that better text representations for questions and narratives can lead to improved performance. Seq2Seq model also did not perform as well. This is due to the inherent difficulty of generation and encoding long sequences. We found that it performs better when training and testing on shorter stories (limited to $30$ statements). Interestingly, the logistic regression baseline performs on a par with MemN2N, but there is still a large performance gap to BiDAF-M and DrQA-M, and the gap is greater for questions that compose multiple relations.

In the \textit{across-world} setting, the performance of all methods dramatically decreases.\footnote{In order to allow generalization across different domains for the Seq2Seq model, we replace entities appearing in each story with an id that correlates to their appearance order. After the model outputs its prediction, the entity ids are converted back to the entity phrase.} This suggests the limitations of these methods in generalizing to unseen relations and vocabulary. The span-based models BiDAF-M and DrQA-M have an advantage in this setting as they can learn to answer questions based on the alignment between the question and the narrative. However, the low performance still suggests their limitations in transferring question answering capabilities.

%Seq2Seq model did not perform as well. One of the reasons that may have caused this large error rate, may be the very long sentence this model receives as input, since it concatenates together all the previous facts. Indeed, when training and testing the Seq2seq model on shorter stories (limited to $30$ facts) it performed much better. 

\section{Conclusion}
In this work, we have taken the first steps towards the task of multi-relational question answering expressed through personal narrative. Our hypothesis is that this task will become increasingly important as users begin to teach personal knowledge about their world to the personal assistants embedded in their devices. This task naturally synthesizes two main branches of question answering research: QA over KBs and QA over free text. One of our main contributions is a collection of diverse datasets that feature rich compositional questions over a dynamic knowledge graph expressed through simulated narrative. Another contribution of our work is a thorough set of experiments and analysis of different types of end-to-end architectures for QA at their ability to answer multi-relational questions of varying degrees of compositionality. Our long-term goal is that both the data and the simulation code we release will inspire and motivate the community to look towards the vision of letting end-users teach our personal assistants about the world around us.

The \textsc{TextWordsQA} dataset and the code can be downloaded at \url{https://igorlabutov.github.io/textworldsqa.github.io/}

\section{Acknowledgments}
This paper was supported in part by Verizon InMind \cite{inmindVision16}. One of the GPUs used in this work was donated by Nvidia.

\bibliography{naaclhlt2018}
\bibliographystyle{naaclhlt2018}

\appendix
\begin{table*}[t!]
\small
\begin{center}
\begin{tabular}{|l@cc|c@cc|c?cc|c?cc|c|}
\hline \bf Dataset & \multicolumn{12}{c|}{\bf Questions } \\ 
\cline{2-13}   \bf & \multicolumn{3}{c@}{Single Entity/Relation} & \multicolumn{9}{c|}{Multiple Entities}\\ 
 & \multicolumn{3}{c@}{} & \multicolumn{3}{c}{Single Relation} & \multicolumn{3}{c}{Two Relations} & \multicolumn{3}{c|}{Three Relations} \\
\cline{2-13}
& $P$ & $R$  & $F_1$ &  $P$ & $R$ & $F_1$ & $P$ & $R$ & $F_1$ & $P$ & $R$ & $F_1$  \\ \cline{1-13}  
\hline \hline \multicolumn{13}{|c|}{\bf Logistic Regression} \\ \hline \hline
\sc Meeting  & 42.0 & 78.1 & 51.0 & 50.6 & 74.6 & 56.6 & 33.3 & 66.3 & 41.1 & 31.8 & 57.6 & 38.0\\
\sc Homework & 39.7 & 57.8 & 44.2 & 98.6 & 99.1 & 98.8 & 57.4 & 78.7 & 62.2 & 25.4 & 42.0 & 28.0\\
\sc Software & 55.0 & 73.3 & 59.0 & 54.3 & 98.2 & 66.5 & 58.2 & 76.0 & 62.3 & 46.3 & 84.6 & 56.4\\
\sc Department & 42.6 & 65.9 & 48.0 & 59.0 & 82.5 & 65.1 & 38.8 & 52.7 & 41.2 & 42.5 & 64.6 & 46.9\\
\sc Shopping & 53.1 & 70.2 & 56.2 & 79.6 & 83.4 & 79.0 & 53.1 & 60.5 & 52.3 & 53.4 & 67.9 & 56.0\\ \hline
Average & 46.5 & 69.1 & 51.7 & 68.4 & 87.6 & 73.2 & 48.2 & 66.8 & 51.8 & 39.9 & 63.3 & 45.1\\
\hline \hline \multicolumn{13}{|c|}{\bf Sequence-to-Sequence} \\ \hline \hline
\sc Meeting & 27.9 & 18.3 & 22.1 & 48.1 & 12.1 & 19.3 & 42.1 & 15.0 & 22.1 & 33.7 & 19.7 & 24.8\\
\sc Homework & 16.3 & 9.0 & 11.6 & 71.9 & 9.3 & 16.4 & 75.3 & 35.9 & 48.6 & 32.9 & 15.6 & 21.1\\
\sc Software & 42.5 & 21.5 & 28.5 & 44.8 & 8.5 & 14.2 & 50.0 & 6.3 & 11.2 & 45.5 & 7.4 & 12.7\\
\sc Department & 49.9 & 35.6 & 41.5 & 54.1 & 20.3 & 29.6 & 57.2 & 38.0 & 45.7 & 43.9 & 39.7 & 41.7\\
\sc Shopping & 25.8 & 16.0 & 19.8 & 71.3 & 28.2 & 40.5 & 33.3 & 19.3 & 24.4 & 46.9 & 31.4 & 37.6\\
\hline
Average & 32.5 & 20.1 & 24.7 & 58.0 & 15.7 & 24.0 & 51.6 & 22.9 & 30.4 & 40.6 & 22.7 & 27.6\\
\hline \hline \multicolumn{13}{|c|}{\bf MemN2N} \\ \hline \hline
\sc Meeting & 56.9 & 56.0 & 54.7 & 66.8 & 58.4 & 58.6 & 57.0 & 57.5 & 54.8 & 38.7 & 40.7 & 38.8  \\
\sc Homework & 42.6 & 41.2 & 41.3 & 97.9 & 63.7 & 73.9 & 60.4 & 47.9 & 49.4 & 36.5 & 29.0 & 30.1  \\
\sc Software & 68.5 & 71.6 & 68.5 & 72.9 & 73.2 & 70.9 & 69.7 & 67.3 & 66.1 & 75.0 & 74.8 & 72.6  \\
\sc Department & 56.3 & 74.3 & 61.3 & 78.5 & 87.0 & 80.2 & 59.4 & 76.6 &
63.2 & 57.8 & 74.2 & 61.6 \\
\sc Shopping & 51.3 & 45.4 & 45.5 & 74.9 & 54.1 & 59.0 & 45.6 & 40.6 &
40.2 & 44.3 & 37.6 & 37.9 \\ \hline 
Average & 55.1 & 57.7 & 54.3 & 78.2 & 67.3 & 68.5 & 58.4 & 58.0 & 54.8 & 50.4 & 51.3 & 48.2 \\
 \hline \hline \multicolumn{13}{|c|}{\bf BIDAF-M} \\ \hline \hline
\sc Meeting & 87.6 & 92.4 & 88.2 & 78.6 & 86.1 & 79.2 & 68.9 & 89.6 & 74.6 & 73.9 & 94.4 & 80.0  \\
\sc Homework & 79.9 & 97.4 & 84.5 & 86.8 & 81.0 & 82.4 & 76.4 & 90.0 & 78.9 & 47.0 & 78.5 & 55.5  \\
\sc Software & 48.0 & 89.4 & 57.4 & 68.5 & 93.6 & 75.8 & 62.4 & 86.1 & 67.5 & 62.7 & 90.9 & 71.3 \\
\sc Department & 57.0 & 64.6 & 58.1 & 73.6 & 85.9 & 76.6 & 67.0 & 83.2 & 70.8 & 63.1 & 71.4 & 64.0  \\
\sc Shopping & 60.5 & 87.1 & 66.9 & 76.7 & 90.9 & 79.8 & 57.1 & 89.0 & 65.8 & 53.2 & 88.5 & 62.0 \\ \hline
Average & 66.6 & 86.2 & 71.0 & 76.8 & 87.5 & 78.8 & 66.4 & 87.6 & 71.5 & 60.0 & 84.7 & 66.6\\ 
\hline \hline \multicolumn{13}{|c|}{\bf DrQA-M} \\ \hline \hline
\sc Meeting & 77.1 & 94.2 & 81.0 & 80.6 & 95.8 & 85.1 & 68.6 & 95.7 & 76.8 & 64.1 & 97.9 & 74.3  \\
\sc Homework & 88.8 & 97.9 & 91.4 & 85.2 & 80.2 & 81.4 & 85.0 & 94.7 & 87.9 & 51.6 & 85.8 & 60.2  \\
\sc Software & 72.7 & 96.0 & 78.9 & 78.6 & 93.3 & 82.7 & 79.4 & 89.4 & 80.9 & 66.3 & 93.2 & 74.5  \\
\sc Department & 67.1 & 97.9 & 76.1 & 80.3 & 95.0 & 84.1 & 67.1 & 94.4 & 74.8 & 55.8 & 95.2 & 66.9  \\
\sc Shopping & 71.5 & 93.9 & 77.7 & 86.4 & 94.8 & 88.7 & 62.8 & 91.1 & 71.4 & 62.4 & 90.7 & 69.7  \\ \hline
Average & 75.4 & 96.0 & 81.0 & 82.2 & 91.8 & 84.4 & 72.6 & 93.1 & 78.4 & 60.0 & 92.6 & 69.1\\
\hline

\end{tabular}
\end{center}
\caption{\label{within-table} Test performance at the task of question answering by question type using the \textit{within-world} evaluation.}
\end{table*}

\begin{table*}[t!]
\begin{center}
\small
\begin{tabular}{|l@c@c|c|c|}
\hline \bf Dataset & \multicolumn{4}{c|}{\bf Questions } \\ 
\cline{2-5}   \bf & Single Entity/Relation & \multicolumn{3}{c|}{Across Entities}\\ \cline{3-5}
 & & Single Relation & Two Relations & Three Relations \\
\cline{2-5}
\hline \hline \multicolumn{5}{|c|}{\bf Logistic Regression} \\ \hline \hline
\sc Meeting  & 8.8 & 10.9 & 7.2 & 5.6 \\
\sc Homework  & 7.5 & 20.2 & 8.5 & 6.7 \\
\sc Software  & 8.2 & 12.0 & 12.9 & 10.6 \\
\sc Department  & 7.4 & 14.4 & 9.7 & 6.1 \\
\sc Shopping  & 8.2 & 9.0 & 5.9 & 6.6 \\ \hline
Average & 8.0 & 13.3 & 8.8 & 7.1 \\
\hline \hline \multicolumn{5}{|c|}{\bf Sequence-to-Sequence} \\ \hline \hline
\sc Meeting  & 7.4 & 8.1 & 10.0 & 14.0 \\
\sc Homework  & 4.2 & 2.9 & 3.1 & 2.3 \\
\sc Software  & 5.0 & 0.6 & 0.9 & 1.1 \\
\sc Department  & 5.5 & 4.0 & 5.6 & 5.6 \\
\sc Shopping  & 2.5 & 2.6 & 2.3 & 2.8 \\
Average & 4.9 & 3.6 & 4.4 & 5.2 \\
\hline \hline \multicolumn{5}{|c|}{\bf MemN2N} \\ \hline \hline
\sc Meeting & 9.0 & 34.2 & 33.0 & 27.4 \\
\sc Homework & 3.3 & 12.4 & 1.0 & 2.5 \\
\sc Software & 13.4 & 0.8 & 3.2 & 2.9 \\
\sc Department & 12.9 & 20.8 & 13.0 & 9.4 \\
\sc Shopping & 0.1 & 0.07 & 0.05 & 0.03 \\
Average & 7.8 & 13.7 & 10.1 & 8.4 \\
\hline \hline \multicolumn{5}{|c|}{\bf BIDAF-M} \\ \hline \hline
\sc Meeting & 31.1 & 40.2 & 30.4 & 30.0 \\
\sc Homework & 10.4 & 20.3 & 2.3 & 7.8 \\
\sc Software & 19.2 & 13.4 & 22.7 & 9.1 \\
\sc Department & 23.3 & 30.5 & 19.0 & 13.5 \\
\sc Shopping  & 5.6 & 3.2 & 2.6 & 3.4 \\ \hline
Average & 17.9 & 21.5 & 15.4 & 12.8 \\
\hline \hline \multicolumn{5}{|c|}{\bf DrQA-M} \\ \hline \hline
\sc Meeting & 44.5 & 58.8 & 33.3 & 37.1 \\
\sc Homework & 19.8 & 30.1 & 5.9 & 9.4 \\
\sc Software & 26.4 & 23.4 & 24.0 & 19.4 \\
\sc Department & 31.0 & 38.8 & 24.4 & 15.7 \\
\sc Shopping  & 19.3 & 2.3 & 6.7 & 7.1 \\ \hline
Average & 28.2 & 30.7 & 18.9 & 17.7\\
\hline
\end{tabular}
\end{center}
\caption{\label{across-table} Test performance ($F_1$ score) at the task of question answering by question type using the \textit{across-world} evaluation.}
\end{table*}

\end{document}